\documentclass[11pt]{article}
\usepackage[margin=1in]{geometry} 
\usepackage{lipsum}
\usepackage{xspace}
\usepackage{array}

\newcolumntype{P}[1]{>{\raggedright\arraybackslash}p{#1}}
\usepackage{amsmath,amssymb,amsfonts}
\usepackage{algorithmic}
\usepackage{graphicx}
\usepackage{subcaption}

\usepackage{array}
\usepackage{algorithm,algorithmic}
\usepackage{hyperref}
\hypersetup{hidelinks=true}
\usepackage{textcomp}
\usepackage{multicol}
\usepackage{authblk}

\usepackage[backend=biber,style=numeric]{biblatex}
\newcommand{\framework}{HALE\xspace}

\date{} 
\title{LLM-powered reasoning in agent-based modeling}

\author{Sifat Afroj Moon\thanks{Corresponding author: moons@ornl.gov}, Dakotah Maguire$^{1}$, Adam Spannaus$^{1}$, Joe Tuccillo$^{2}$, Maksudul Alam$^{3}$, Sudip K. Seal$^{3}$, John Gounley$^{1}$ and Heidi Hanson$^{1}$}
\affil{$^{1}$Computational Science and Engineering Division, Oak Ridge National Laboratory, Oak Ridge, TN 37830, USA\\
$^{2}$Geospatial Science and Human Security Division, Oak Ridge National Laboratory, Oak Ridge, TN 37830, USA\\
$^{3}$Computer Science and Mathematics Division, Oak Ridge National Laboratory, Oak Ridge, TN 37830, USA}

\begin{document}
\maketitle
\begin{abstract}

Agent-based modeling (ABM) has the capability to model millions of individual humans and their interactions, which is valuable for constructing a population-scale digital twin to support policy decisions.
However, ABMs have traditionally relied on static and outdated data, which prevents the models from adapting to real-time changes in a given scenario.
Complete real-time information about the activities of a population is often dynamic, as mobility depends on epidemic conditions and human decision-making.
The research presented here provides a novel approach to addressing this information gap.
Large language models~(LLMs) offer new opportunities to predict human decision-making as a scenario unfolds.
To this end, we introduce a scalable Hybrid Agent-based and Language-driven Epidemic (\framework) modeling framework that leverages LLMs to predict human decision-making in an ABM simulation.
As a proof-of-concept, we use \framework to simulate COVID-19 and its effects in Salt Lake County, UT, from September 2020 to February 2022.
The \framework framework better captures the observed epidemic peak and size, whereas simulations that use only ABM tend to overestimate the epidemic's effects.

\end{abstract}


\section{Introduction}

\label{sec:introduction}
{A}{gent-based} modeling (ABM) is a powerful tool for modeling complex dynamic systems at the individual level.
Understanding the impact of individual behavior on disease spread is critical for accurate disease forecasts.
ABM enables us to model individual human behavior (e.g., mask-wearing, vaccine acceptance) based on a distribution of a population, their movement over time, and their social interactions~\cite{afroj2023all}.
Typically, ABMs are initialized with these parameters, and the complex interactions of agents are modeled over a set time period.
However, human behavior is inherently uncertain and depends on various factors, including social environment, geographic location, climate, and weather.
Human preferences also change over time based on external environments and social pressures.
The reasoning abilities of large language models~(LLMs) can assist in modeling change in human behavior during a disease outbreak~\cite{chopraagenttorch,zhang2025llm, vezhnevets2023generative, ghaffarzadegan2023generative}.

Although advancements in LLMs have improved knowledge across various domains~\cite{zhang2025exploring}, their application in epidemic modeling---especially individual-based network modeling---has not yet been fully explored.

LLMs can assist epidemic modeling and simulation in various ways, such as performing literature reviews, 
forecasting epidemics~\cite{kwok2024utilizing}, and predicting human reactions to an epidemic~\cite{chopraagenttorch}.
LLM agents can also exhibit realistic human behaviors~\cite{ferraro2024agent}.
Interestingly, the collective intelligence of a generative agent-based model can mimic real-world responses to disease spreading~\cite{williams2023epidemic}.
In this study, we investigate how LLMs can infer the change in human behavior during an epidemic and support ABMs in dynamically updating network structures.
These ABM-based population-scale digital twins use real-world human activity data, which consist of different types of activities, including staying at home, going to school, being in the office, and shopping, along with their respective locations.
However, detailed activity data are often difficult to obtain, and this difficulty often requires researchers to generate network models from historical survey and demographic data.
Moreover, these models do not account for the change in an individual's activities based on perceived threat or social pressures.
However, LLMs may be useful for modeling changes in behavior as the perceived threat level evolves over time.

Network development is one of the pre-processing steps in individual-based network simulation.
In these networks, nodes can represent different entities (e.g., people, locations, animals, plants) depending on the research objective.
In epidemic models for diseases such as COVID-19, nodes represent individuals, and links represent connections between two people~\cite{ moon2021contact}.
These links form when two people come into close proximity or visit the same location at the same time.
Information on individuals' activities, including time and location, can be used to construct an activity-based temporal network.
However, detailed activity data are often missing or only partially available.
Moreover, activity data may be available for a population during a specific time period, whereas inference is required for another period.
In an individual-based temporal network model, the activities can be divided into two categories: deliberate and non-deliberate actions.
LLMs can assist ABMs in modeling these deliberate actions when creditable data are not available.
For example, in an epidemic ABM simulation, spreading a virus from an infected person is a non-deliberate action, whereas going out for outdoor activities is a deliberate action of an agent.
Deliberate actions indirectly influence non-deliberate actions.
Additionally, agents in different locations may make different choices under similar conditions; they may perceive threats differently than their neighbors and react accordingly.
In this research, we address these data gaps by leveraging LLMs within ABM simulations.

In this research, we have developed a spatiotemporal Hybrid Agent-based and Language-driven Epidemic (\framework) framework for near real--time forecasting.
LLM-powered ABM simulation is a promising new research area. However, most of the research on this topic is in other domains, is conceptual in nature, and has not yet reached the development phase~\cite{liu2025agentic}.
Researches have also replaced ABM agents with generative agents and modeled their behavior with generative AI.
Unfortunately, this approach is often computationally expensive and not intuitive or scalable~\cite{williams2023epidemic}.
Therefore, we have developed a hybrid framework that leverages the strengths of both ABM and LLMs.
The \framework framework can simulate ABM models with millions of agents while enabling computationally efficient LLM inference.

The major contributions of this research are as follows: 
\begin{itemize}
    \item A scalable workflow to construct dynamic temporal networks models using a partially available real-world activity dataset comprising millions of agents.
    
    \item A hybrid agent-based and language-driven epidemic modeling framework (\framework) to enable near real-time forecasting.
    Previous studies have primarily used ABM for epidemic analysis from retrospective data.
    Although these methods are capable of forecasting, they do not adapt to real-time changes in external environments and perceived threat level.
    In reality, however, an individual's mobility varies dynamically in response to an evolving epidemic situation.
    In our \framework framework, we incorporate an LLM as a feedback loop to adjust human mobility in response to epidemic conditions.
    
    \item The performance analysis of the \framework framework by using an ablation study.
    We also provide an analysis of the LLM's inference across different spatial population groups.
\end{itemize}


ABM is an important decision tool as it can model effects of different public health decisions (e.g., lockdown, mandatory mask-wearing) at the individual level.
However, people often make different decisions depending on their surroundings and demographics features, which ABM cannot capture.
To address this challenge, we develop the \framework framework in which LLMs assist ABM model in individual-level decision-making. The LLM supports the \framework framework to adjust mobility in response to an outbreak. The research observes that different population groups react differently to an outbreak depending on their demographic characteristics.

\section{Methods}
\subsection{Datasets}\label{subsec:data}
To develop the ABM, we use population data for Salt Lake County, UT, which represents a population of approximately 1.1 million.
We then use Oak Ridge National Laboratory’s UrbanPop model to generate the synthetic population~\cite{tuccillo2023urbanpop}.
Each individual in the synthetic population set has the following features assigned: age, race, sex, home ID, geographic location (latitude, longitude), living arrangement, occupation, and income level.
The statistical distribution of the synthetic population matches with the American Community Survey's Public-Use Microdata Sample (PUMS) at the census block group level.

Each individual is assigned to a specific activity for a specific time by using data from the 2017 National Household Travel Survey (NHTS)~\cite{NHTS}.
The Federal Highway Administration conducts the NHTS, which serves as an authoritative source of information on the travel behavior of the US public and enables analysis of trends in personal and household travel.
The activity schedule data include activities for each individual for every hour of the week, and an individual may engage in the same activity for multiple consecutive hours.
The activity schedule data contain information on time, activity type, home location, and activity location.
Locations are provided using Uber H3 hexagonal cells at a level-8 resolution~\cite{H3}.

The UrbanPop synthetic population includes nighttime (residential) and daytime (work or school) locations at the census block group level.
Each individual is assigned a residential point location (house) from FEMA's USA Structures dataset ~\cite{yang2024baseline} by matching synthetic household dwelling characteristics with the properties of residential structures available within each block group~\cite{tuccillo2024downscaling}.

Table~\ref{tab:schedule} presents a part of the activity schedule data for two persons.
In the synthetic population, each person has a unique identifier, $PID$.
Each $PID$ is linked to a household ID, $HID$, which is a residence located in a hex H3 (see \textit{Home assignment} column in Table~\ref{tab:schedule}).
Agent $PID_{01}$ in Table~\ref{tab:schedule} is at home on Monday from 0600 to 0700.
Therefore, the H3 hex IDs for the home location and the activity location are the same for this activity.
This individual (agent) goes to school after 0700, and the school location is in H3\_hex\_id 5.
Similarly, $PID_{02}$ has a work activity from 0700 to 0800, with the work location in H3\_hex\_id 8.
The activity schedule dataset contains approximately 189 million entries covering 168 hours ($7\times24$) for a population of 1.1 million in Salt Lake County, UT.

The activity schedule data include 22 distinct activities.
The activity list $A$ is \{home, school, work, exercise, services, retail, leisure, religious/community, meals, visiting friends and relatives, drop-off/pickup, travel, changing transportation mode, errands, volunteering, medical, childcare, adult care, work home, work related, other, and unknown (activity purpose not given by respondent)\}.
The activities for home, school, and work are anchored activities in the schedule.
The activity schedule data contain activity location information only for the anchored activities (last column of the Table~\ref{tab:schedule}).

\begin{table*}[htpb]
\centering
\caption{Sample of activity schedule data at the individual level.}
\begin{tabular}{ |P{1.5cm}|P{1cm}|P{2cm}|P{1.3cm}|P{2.7cm}|P{2.5cm}|P{2.7cm}|} 
\hline
\textbf{Person ID, $PID$} & \textbf{Day} & \textbf{Hour} & \textbf{Activity} & \textbf{Home location (H3 hexagonal spatial index)} & \textbf{Home assignment, $HID$ }& \textbf{Activity location (H3 hexagonal spatial index)}\\ 
\hline
\hline
$PID_{01}$ & Monday & 0600--0700& Home & H3\_hex\_id 1& H3\_hex\_id 1\_7 & H3\_hex\_id 1\\
$PID_{01}$ & Monday & 0700--0800& School & H3\_hex\_id 1& H3\_hex\_id 1\_7 & H3\_hex\_id 5\\
$PID_{01}$ & Monday & 0800--0900& School & H3\_hex\_id 1& H3\_hex\_id 1\_7 & H3\_hex\_id 5\\
$PID_{02}$ & Tuesday & 0600--0700& Home & H3\_hex\_id 1 &  H3\_hex\_id 1\_10 & H3\_hex\_id 1\\
$PID_{02}$ & Tuesday & 0700--0800& Office & H3\_hex\_id 1 &  H3\_hex\_id 1\_10& H3\_hex\_id 8\\
\hline
\end{tabular}
\label{tab:schedule}
\end{table*}

\subsection{Activity-based temporal networks development} \label{subsec:network_dev}
We develop a network, $G_{a}(t)$, for each activity at each hour. Therefore, for one week, we have 3,696 $G_{a}(t)$ networks that correspond to 22 activities. Each network consists of 1,133,023 nodes (agents).
The network development algorithms vary depending on the activity.

\subsubsection{Hourly networks for anchored activities}
\paragraph{\textbf{Home activity:}}
The \textit{home} activity contains point locations for each residential place.
At hour $t$, agents or $PID$s that share the same $HID$ and have the activity \textit{home} or \textit{work-home} are connected in the network $G_a(t)$ for the \textit{home} activity at hour~$t$.

\paragraph{\textbf{School activity:}}    
The \textit{school} activities also contain school locations.
From the activity schedule dataset, we first calculate the total number of students for each school.
Some schools have few students (fewer than 19), whereas others have many (more than 500).
In the hourly school activity network $G_a(t)$, students from smaller schools (fewer than 19 students) are fully connected if they participate in the \textit{school} activity at time $t$.
    
We divide students from larger schools into classes and maintain the average public school class size based on data from the National Center for Education Statistics~\cite{NCES}.
We keep a record or map that links students to their respective classes to ensure that each student attends the same class every day.

For the \textit{school} activity hourly network, we create two types of networks: the in-class network and the out-of-class network.
In the in-class network, students belonging to the same class and having the \textit{school} activity at time $t$ form a fully connected network.
For the out-of-class network, we construct an Erd\H{o}s--R\'enyi random network among students from the same school, with an average degree of 4 within the 8-hour period~\cite{del2007mixing}. We assume that students have eight hours of school each day.
The out-of-class network represents contacts outside of the student’s own class (e.g., gym, cafeteria, school bus). Finally, we merge the in-class and out-of-class networks to form the complete hourly \textit{school} activity network.
    
\paragraph{\textbf{Office activity:}} The \textit{office} activities contain information about the office building location at the H3 level 8 index ($\sim$0.74~km$^2$); the exact office is unknown.
Within each hex, some office buildings have many workers, whereas others have few, similar to schools.
However, the data do not include the exact office location. They also do not include information about the layout, such as whether it is open-plan or composed of cubicles.
Therefore, we do not divide offices into smaller units and utilize a configuration random network model to estimate contact.
In the hourly \textit{office} activity network, we create a configuration random network in which the degree of each node is assigned from a Gaussian distribution: $\mathcal{N}(\mu, \sigma^2)$.
For the age range of 18--59, we consider hourly $\mu = \frac{21.154}{8}$ and $\sigma^2 = \frac{10.58}{8}$~\cite{del2007mixing, moon2021contact}. We assume that the workday is eight hours.

\subsubsection{Hourly network for non-anchored activities}
The \textbf{non-anchored activities} (e.g., meals, retail, exercise) do not have exact activity location information.
Therefore, we create contacts for them differently.
We use a spatial network model~\cite{moon2019spatio, mundt2009long}, specifically an exponential distance dispersal kernel network, to construct hourly contact networks for each \textit{non-anchored activity}.
In this spatial network model, the probability of a connection between two nodes decays exponentially with distance.

We select the spatial network model to capture the phenomenon in which people from nearby locations have a higher probability of meeting each other because they may go to the same restaurant, shop, or gym given their proximity.
In the exponential distance kernel, the probability, $P(i,j)$, forms a link between node $i$ and $j$:
\begin{equation}
    P(i,j) = \lambda e^{-\lambda d_{ij}} 
\end{equation}
Here, $\lambda$ is the average travel distance for an activity, and $d_{ij}$ is the distance between the home locations of $i$ and $j$.
Average travel distance $\lambda$ varies across different activities.
In our spatial network model, we select $\lambda$ values for the United States based on various previous surveys, research studies, and reports.
For example, the average travel distance for exercise is 4~miles~\cite{WSJ}; average commuting time to church is around 15~minutes and roughly 11.25~miles; the average distance between food establishments and homes is 2.6~miles~\cite{liu2015beyond}; average travel distance for childcare is approximately 5~miles~\cite{blumenberg2024decisions}; and a 2017 study reports the average distance for medical/dental care travel in the United States is 10.04~miles~\cite{akinlotan2023travel}. For activities for which average travel distances are unknown, we use the mean value of the known $\lambda$s.

\subsubsection{Hourly networks for all activities}
For each hour, we develop 22 networks that correspond to 22 different activities.
To create the hourly network, we merge all the individual activity networks.
The hourly network contains no duplicate links because the activity schedule data ensure that a person or agent cannot be in two locations simultaneously.
The edge list of the hourly network records the activity associated with each edge as an attribute.
The average degree of the hourly networks differs between daytime hours 
and nighttime hours.

\subsubsection{Daily networks}
We merge the hourly networks to create daily networks.
In the daily networks, we add an additional edge attribute called \textit{weight}.
If an edge appears in $h$ hourly networks for the same activity, then its weight in the daily network is set to $h$. Table \ref{tab:adja_list} shows a portion of the adjacency list of a daily network. The first row in Table \ref{tab:adja_list} indicates that $PID_{01}$ and $PID_{05}$ are connected in the daily network, with a weight corresponding to five hours of office activity in the same location. 

These daily networks are used for the ABM simulation.
The average number of connections per agent in the daily networks is approximately $22.41$, which is consistent with previous literature~\cite{del2007mixing}.
\begin{table}[htpb]
\centering
\caption{A portion of the adjacency list of a daily network.}
\begin{tabular}{ |P{1.3cm}|P{1.3cm}|P{1cm}|P{1cm}|P{1.5cm}|} 
\hline
\textbf{Person ID, $PID$} & \textbf{Person ID, $PID$} & \textbf{Day} & \textbf{Weight} & \textbf{Activity} \\ 
\hline
\hline
$PID_{01}$ & $PID_{05}$ & Mon& 5 & Office\\
$PID_{01}$ & $PID_{05}$ & Mon& 1 & Exercise \\
$PID_{01}$ & $PID_{03}$ & Mon& 0.5 & Meals \\
$PID_{02}$ & $PID_{10}$ & Mon& 6 & School \\

\hline
\end{tabular}
\label{tab:adja_list}
\end{table}
\subsection{The \framework Framework}\label{sec:framework}
\subsubsection{Architecture}
\framework is an individual-based framework that couples dynamic, AI-generated mobility patterns with ABM.
In \framework, $N$ agents (or individuals) are connected via an activity-based temporal network, $G(t)$.
Each agent, $n_i$, is associated with a property vector, $P$.
The properties can be demographic information, including sex, age, home location, school, workplace, average travel distance, and education.

\begin{figure}[htbp]
\centering
\includegraphics[width=0.99\textwidth]{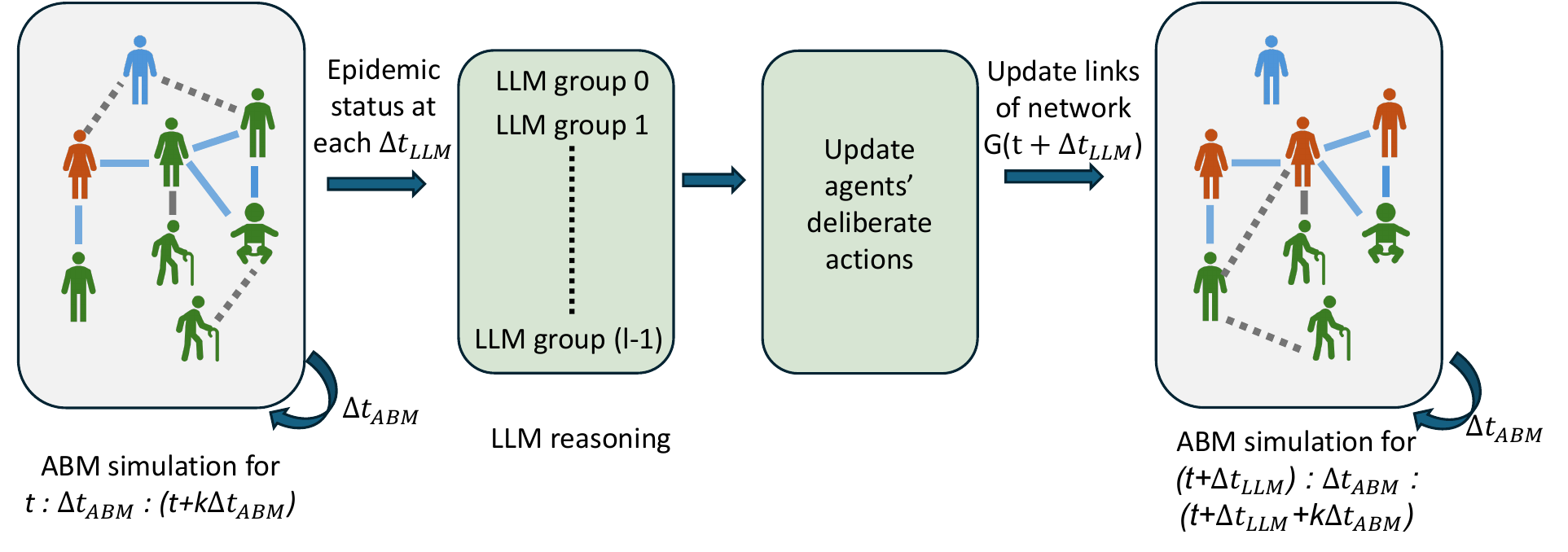}
\caption{The \framework framework—scalable agent-based simulation with LLM reasoning—is applied to an individual-level network model.
The LLM updates specific deliberate actions of agents.
In this model, agents are connected via temporal, activity-based links.
Solid edges represent connections formed by fixed actions. These connections are temporal but do not depend on LLM reasoning—for example, connections formed by living in the same home. 
Dotted edges indicate connections resulting from deliberate actions that evolve with LLM reasoning.
The colors of individuals---green, orange, and blue---represent different health states. ABM updates the health state of agents after each ABM time step, $\Delta t_{ABM}$. After each $k\Delta t_{ABM}$ (or $\Delta t_{LLM}$) time step, network updates the dotted edges from LLM reasoning. }
\label{fig:framework}
\end{figure}

Figure~\ref{fig:framework} presents the architecture of the \framework framework.
The ABM component of the framework models epidemic spreading by using a disease model such as susceptible-infected-susceptible (SIS), susceptible-infected-recovered (SIR), or susceptible-exposed-infected-recovered (SEIR).
A health status vector, $H\in\mathbf{R}^{N}$, tracks the health status of all agents.
ABM updates the health states of agents at discrete time steps, $\Delta t_{ABM}$.
For the SIR disease model, an agent's health state is $h_i\in\{S, I, R\}$.
At each time step, $\Delta t_{ABM}$, infected agents can spread the disease to their healthy neighbors in the network $G(t)$, with a transmissibility, $\beta$.
Then, the state of a healthy agent changes to infected, $h_i: S \rightarrow I$.
These events depend on the network or connectivity.
Infected agents recover, $h_i: I \rightarrow R$, with a probability, $\gamma$. This event occurs independently of the network structure.

For the LLM component, the agents are divided into $L$ subpopulations or groups based on their demographic properties and residential locations.
Here, $L = \{l_1, l_2, l_3,......\}$, $l_j = \{n_i, n_{i+1}...\}$, and $|N| \gg |L|$.
At $t=0$, we initialize $L$ LLM agents -- one for each group.
The groups or LLM agents have different sizes, $|l_i| \neq |l_j|$.
LLM agents are distinct and independent.
The outputs of the LLM agents are used in ABM simulation.
Therefore, we use structured LLM output formats~\cite{geng2025jsonschemabench} to make the LLM inference machine-readable.

The ABM simulation then models the disease dynamics of $N$ agents on the network $G(t)$ as time evolves.
The LLM agents generate reasoning for mobility patterns of the $L$ LLM agents based on the real and perceived environment over time.
Then, the \framework framework will update the network, $G(t)$, based on the LLM reasoning.
For example, if an LLM agent $l_i$ decides to decline an outdoor activity and stay at home, then links formed from that activity for the subpopulation for that LLM agent $l_i$ in $G(t)$ will be deactivated and will not spread the disease for that time.
ABM simulation updates the states of agents at each ABM time step, $\Delta t_{ABM}$, while the LLM assesses the current state of the system and provides inference on mobility at each LLM time step, $\Delta t_{LLM}$.
After each $\Delta t_{LLM}$ time step, ABM simulation waits for the LLM inference.
Because the ABM requires more frequent updates compared to the LLM component in the \framework framework, we have $\Delta t_{LLM} = k\Delta t_{ABM}$, where $k$ is an integer greater than 1.

\subsubsection{Spatial LLM grouping}\label{subsec:llm_group}
\begin{figure}[htpb]
\includegraphics[width=0.9\textwidth]{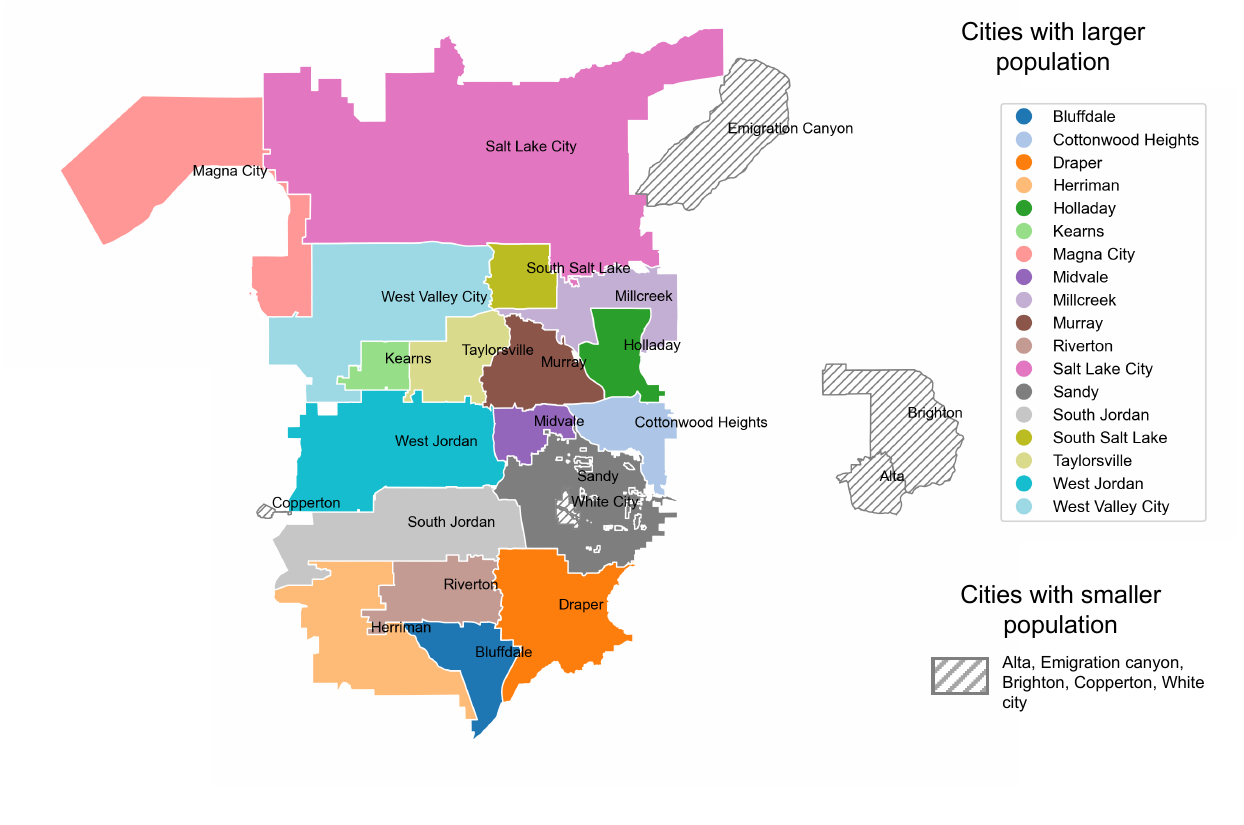}
\caption{Spatial LLM grouping for Salt Lake County, UT.
We use race, sex, and age to further subdivide the people in cities with larger populations.}
\label{fig:SaltLake}
\end{figure}

The ABM simulation contains millions of agents.
To add LLM reasoning to an ABM simulation while maintaining a reasonable compute time and  accounting for Tobler's First Law of Geography, we group the agents based on their spatial location and demographic properties.
We consider 23 municipalities in Salt Lake County, UT: Bluffdale, Cottonwood Heights, Draper, Herriman, Holladay, Kearns, Magna City, Midvale, Millcreek, Murray, Riverton, Salt Lake City, Sandy, South Jordan, South Salt Lake, Taylorsville, West Jordan, West Valley City, Alta, Brighton, Copperton, Emigration Canyon, and White City (Figure~\ref{fig:SaltLake}).

We divide the municipalities into two categories based on population size: larger cities and smaller cities.
Alta, Brighton, Copperton, Emigration Canyon, and White City are classified as smaller cities, whereas the remaining municipalities are classified as larger cities.

We further divide the population of the larger cities by sex \{male, female\}, race \{White, Black or African American, Asian, Native American, Pacific Islander, Multiethnic, Other\}, and age group \{0--18, 19--24, 25--34, 35--44, 45--59, over 60\}.
However, we do not further divide the population of the smaller cities because this would yield very small, often empty, groups.
In total, there are 1,552 population groups, each represented by a corresponding LLM agent.

\subsubsection{ABM-LLM structure}\label{subsec:abm_llm}
In this hybrid setup, the ABM and LLM run sequentially, passing their results back and forth to each other. Here, ABM simulation and the LLM have different time scales: hourly time steps $\Delta t_{ABM}$ for ABM simulation and weekly time steps $\Delta t_{LLM}$ for the LLM.
These time steps are flexible and can be different for different research objectives.

\paragraph{\textbf{ABM simulation:}}
ABM will simulate the disease progression in the daily weighted network, $G(t)$, where, $t\in\{day 0, day 1, day 2,...\}$ (Algorithm 1 in \cite{spannaus2025data}).
For our experiment, we choose the SIR disease model and use incidence data for COVID-19.

We design the ABM simulation to model disease spreading using the Repast modeling framework~\cite{ozik2021population, spannaus2025data}.
Repast is a highly scalable distributed framework that can leverage multiple processing cores via MPI (Message Passing Interface).

\paragraph{\textbf{LLM reasoning:}} In this research, we use LLMs to explore an important question: whether a person would continue their normal daily activities (e.g., school, work, shopping) if they were aware that a disease is spreading in their community.

Here, the LLM takes the incidence level from ABM and provides reasoning for different LLM groups.
After each $\Delta t_{LLM}$, ABM sends the simulation or disease scenario to the LLM, which then provides the reasoning for each LLM agents. 

We use zero-shot learning and assume that the LLM already have the baseline knowledge needed to infer typical human activities in a given area~\cite{civitarese2025large}.
For our experiments, we use pretrained and instruction-tuned generative models with 8 billion parameters from Llama 3.1~\cite{dubey2024llama}. We also utilize vLLM, an open-source library for fast inference of LLMs~\cite{kwon2023efficient}.
In the configuration object of the vLLM, we use $temperature = 0.2$, a balanced setting for creative yet coherent outputs.
To increase efficiency in the hybrid system, we use batch processing for the LLM agents.
    
Because the output of the LLM has been used as feedback for ABM, it is essential to generate structured output from the LLM~\cite{geng2025jsonschemabench}.
We use the recently developed Outlines framework to obtain structured output in a yes/no format~\cite{willard2023efficient}. If the LLM output of an LLM agent $l_i$ is \textit{yes} for a week $w$ for a certain activity, then the corresponding ABM agents contacts in the network $G(t)$ will remain unchanged.
If the LLM output of an LLM agent $l_i$ is \textit{no} for a week, $w$, for a certain activity, then the corresponding ABM agents contacts in the network $G(t)$ will be deactivated.
Therefore, an infected agent $n_i$ cannot infect its susceptible neighbors because
$n_i$ will no longer have connections with them, as it will stay home. For $\Delta t_{LLM}$, we take week (seven days) as the observation data from CDC is weekly for COVID-19~\cite{Cramer2021-hub-dataset}. However, $\Delta t_{LLM}$ is flexible and can take any value depending on the research question.
 
\paragraph{\textbf{LLM-ABM server-client integration on an HPC platform:}}
We integrate the LLM as a GPU-backed 
inference server and ABMs as clients. The LLM runs as a single-process, single-GPU inference server.
The communication between the LLM and ABM is performed by FastAPI, Uvicorn, and python library requests.
The framework runs separate Slurm jobs for the LLM and ABM. 
At each LLM time step $\Delta t_{LLM}$, ABM sends an update and query to the LLM server for the inference. The ABM pauses the simulation for the LLM inference. After receiving a response from the LLM server, it updates G(t) and resumes the ABM simulation.
Notably, one server can respond to multiple ABM clients at a time.

\section{Results and Discussion}\label{sec:results}
In this section, we describe the \framework framework applied to a real-world epidemic scenario: COVID-19 in Salt Lake County, UT. To perform our experiments, we use one compute node from the Frontier HPE Cray EX supercomputer hosted at Oak Ridge National Laboratory. This compute node features AMD MI250X GPUs.
 
Our experiments simulate the COVID-19 outbreak from September 2021 to February 2022. We use an individual-based SIR epidemic model. We adopt the average basic reproduction number and recovery rate that correspond to the Omicron variant. Previous studies report that the average basic reproduction number for the Omicron variant is 9.5 (inter quartile range: 7.25, 11.88)~\cite{liu2022effective}, with a recovery rate of $\frac{1}{5} days^{-1}$~\cite{ukhsa2023covid}.
Because we use an individual-based network framework, we adjust the transmissibility for each edge according to the average degree.
 
\subsection{\framework framework simulation}
\begin{figure}
\centering
\begin{subfigure}{0.49\textwidth}
    \centering
    \includegraphics[width=\linewidth]{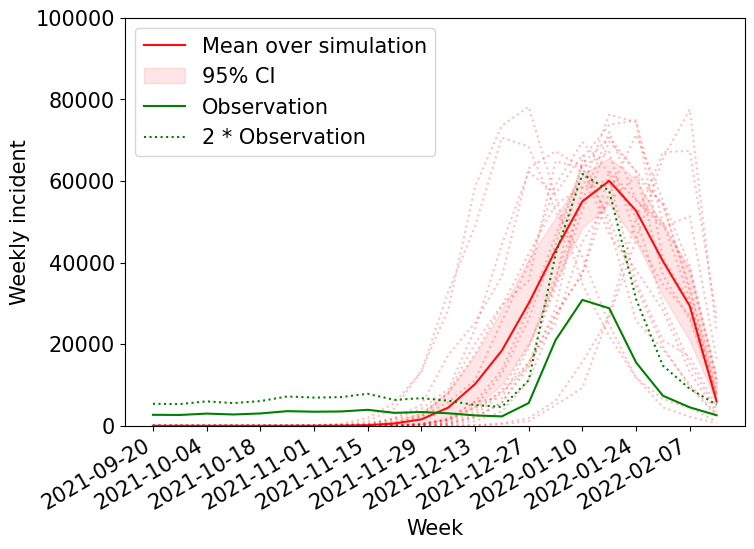}
    \caption{\framework}
\end{subfigure}
\hfill
\begin{subfigure}{0.49\textwidth}
    \centering
    \includegraphics[width=\linewidth]{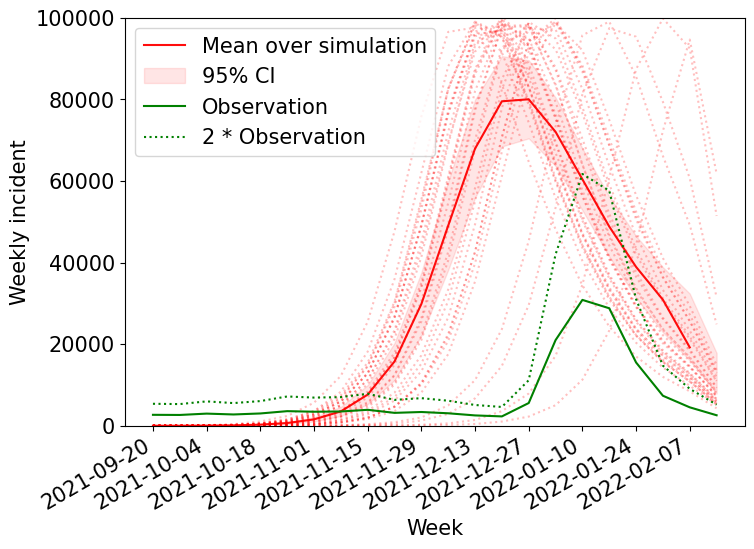}
    \caption{ABM-only}
\end{subfigure}
\caption{Weekly new COVID-19 cases from September 2021 to February 2022 in Salt Lake County, UT. The green line shows weekly case data from the Centers for Disease Control and Prevention. The red lines represent simulation outputs. Subfigure (a) shows simulation results from the \framework framework, and subfigure (b) shows simulation results from the ABM. The red-dotted lines represent individual simulations, and the red-solid line represents the mean of all simulations.}
\label{fig:weekly_inci}
\end{figure}

As the \framework framework is stochastic by nature, we simulate it for 30 times and compare the aggregated results with real-world observations. 
Figure~\ref{fig:weekly_inci} shows new weekly case data.
The red-dotted lines represent individual simulations, and the red-solid line represents the mean of all simulations.
The green line shows weekly COVID-19 case data in Salt Lake County, UT~\cite{Cramer2021-hub-dataset}.
The peak time of the mean over simulation line follows the observation.
However, the area under this red curve or the total number of infections in the experiments is 2.21 times higher than the observation (green line).
Because the observation data mostly included the symptomatic cases and missed many asymptomatic cases, it is lower than the actual infected population.
Wang et al. found that asymptomatic population is almost 44\% of the total infection~\cite{wang2023asymptomatic}.
Another study estimated that presymptomatic and asymptomatic cases together comprise at least 50\% of the force of infection at the outbreak's peak~\cite{subramanian2021quantifying}, which is consistent with our simulation results.

Figure~\ref{fig:weekly_inci}b presents the simulation output from ABM-only with the same epidemic model, parameters, and contact networks. Within the \framework framework, the average number of links decreases as people choose to stay at home rather than go out in response to increasing disease levels.
Therefore, for a fair comparison, we randomly deactivate 35\% of contacts in the ABM-only simulation.
The ABM-only simulation still fails to follow the actual observation in terms of peak timing and epidemic size.

\subsection{LLM inference}
We use the offline Llama model for behavioral prediction of the LLM agents. Our observations find that LLM reasoning is sensitive to prompts and the temperature hyperparameter. We use a temperature setting for the Llama model that balances randomness and determinism to avoid overly random or overly deterministic outputs.
Very Low temperature settings act deterministically and introduce very little randomness when generating the next token.
At $temperature = 0$, the model behaves very conservatively, and the probability of declining outdoor activities for all age groups is almost always close to 1.
Higher temperatures introduce more randomness in the inference and reduce meaningful patterns in the generation.
For example, at $temperature = 0.7$, the probability for all groups at any time is approximately 0.5.
Therefore, our experiment uses a temperature of 0.2. 

A sample prompt for an LLM agent, $l_i$, in the experiment is provided below.

\vspace{1em} 
\begin{center}

\fbox{\parbox{0.7\linewidth}{%
\begin{tabular}{p{1.1cm} p{9cm}}

\textbf{Prompt:} & \textbf{Behavior prediction task} \\
Location: & Salt Lake County, UT  \\
Context:   & 0.02\% people infected by COVID-19 \\ 
Person: & Asian female, age 25--34, lives in Bluffdale, Salt Lake County, UT \\
Question: & Considering the infected percentage and a person’s behavior based on demographic factors, will this person go to a public place?
\end{tabular}}%
}
\end{center}
\vspace{1em}
We calculate the probability of declining an outdoor activity by LLM agents from the HALE simulations.
\subsubsection{Analysis of LLM agents across different age groups over time}
\begin{figure*}[htpb]
\includegraphics[width=0.99\textwidth]{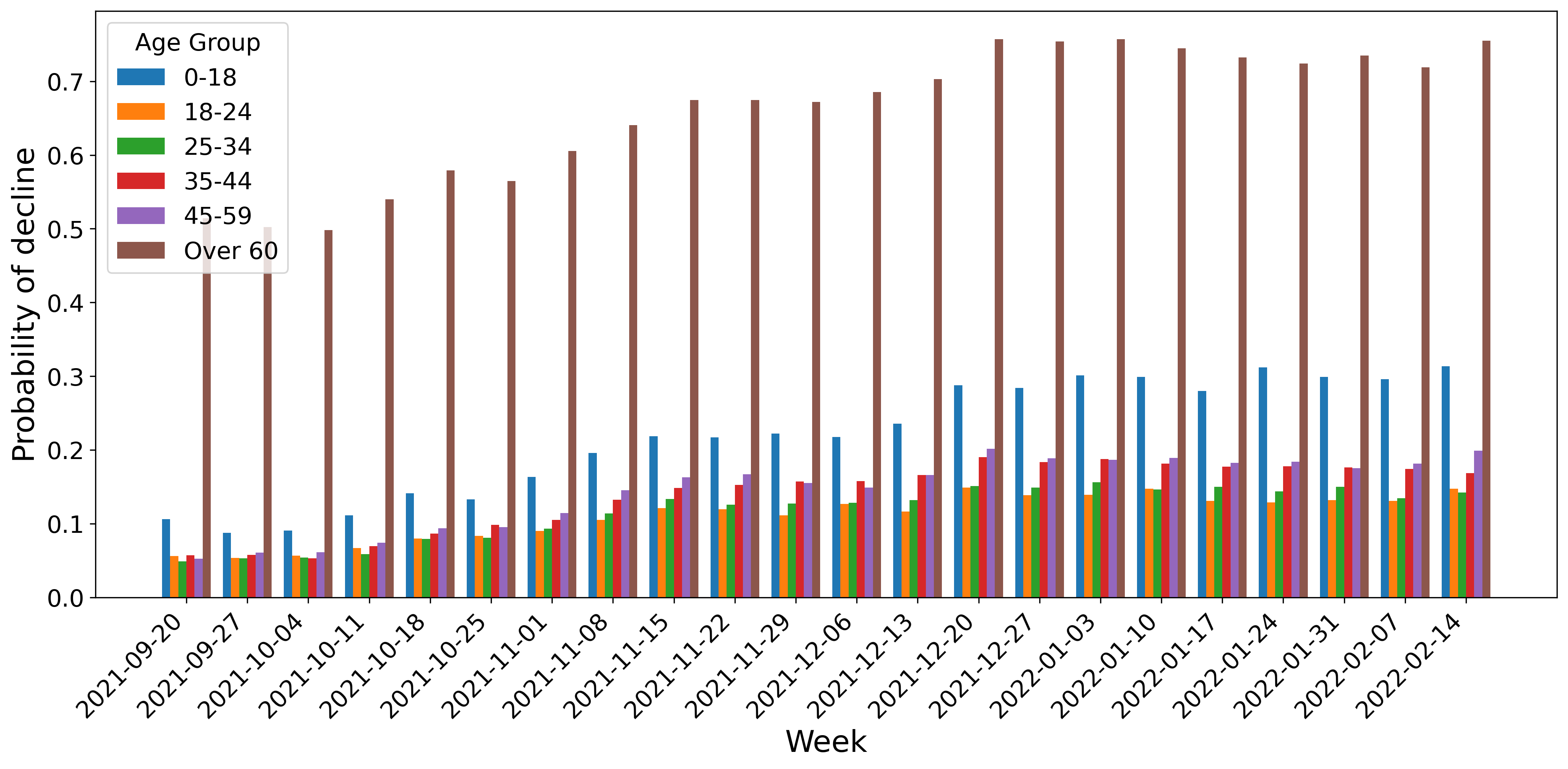}
\caption{LLM inference for different age groups in the \framework framework. The $y$-axis shows the probability to decline participation in an outdoor activity.}
\label{fig:LLM_age}
\end{figure*}

Figure~\ref{fig:LLM_age} shows the probability of declining an outdoor activity with time for different age groups.
The probability of declining outdoor activities increases for all groups over time as the number of infected people increases (Figure \ref{fig:weekly_inci}).
The over 60 and 0--18 age groups consistently exhibit higher probabilities, whereas the 18--24 and 25--34 age groups show lower probabilities among all groups.
The LLM inference behaves as expected for age group analysis.

Experiments show that the probability of declining an outdoor activity for all age groups increases slowly and does not exhibit any sharp peak, unlike the weekly infected cases shown in Figure~\ref{fig:weekly_inci}. This indicates that human behavior changes dynamically in response to surroundings, but the changes occur gradually.

\subsubsection{Analysis of LLM agents across different municipalities}
\begin{figure*}[htpb]
\includegraphics[width=0.99\textwidth]{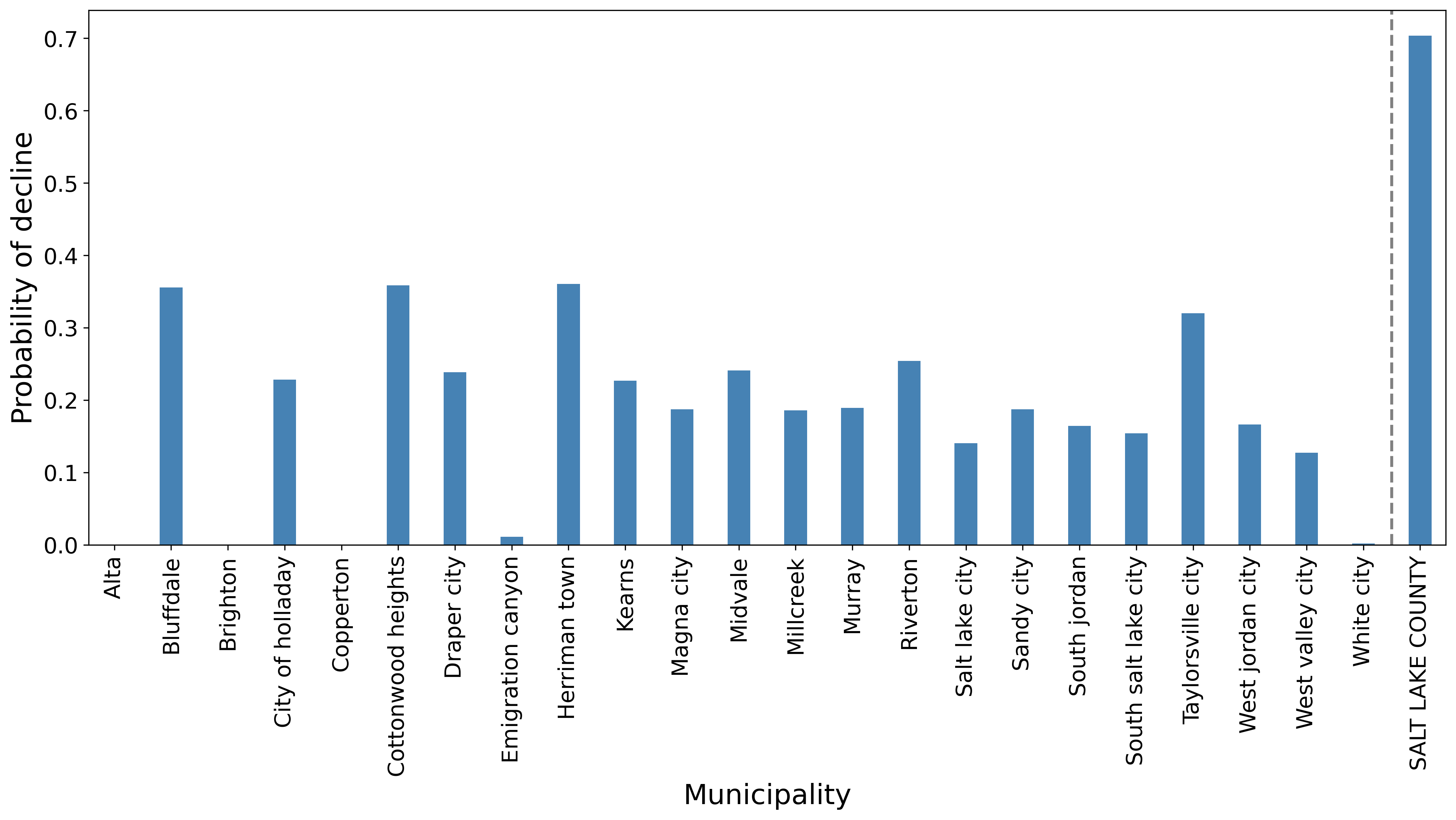}
\caption{LLM inference in the \framework framework across different municipalities. The $y$-axis shows the probability to decline participation in an outdoor activity.}
\label{fig:LLM_city}
\end{figure*}

In the analysis of LLM inference (Figure~\ref{fig:LLM_city}), residents of different municipalities or cities show different levels of response to an epidemic.
Figure~\ref{fig:LLM_city} also shows that information referring only to the county, without specifying a city, is associated with a higher probability of declining outdoor activity. In the experiments, agents living in Salt Lake County, UT, without any city-specific information have a probability of declining outdoor activity of around 0.7 (last bar in the Figure~\ref{fig:LLM_city}).
However, when the city of the LLM agent within Salt Lake County is specified, the probability is lower and takes different values. 

In the experiment, Bluffdale, Cottonwood Heights, Herriman town, and Taylorsville city have a higher probability of decline. Bluffdale and Harriman are rapidly growing suburbs with many young families. These cities also have a higher proportion of children (under 18) compared to the Utah state average. On the other hand, Cottonwood Heights has a relatively older population on average. Due to their population structure and specfic locations, these cities have a higher probability of declining an outdoor activity during an outbreak.

\begin{figure}[htpb]\centering
\includegraphics[width=0.35\linewidth]{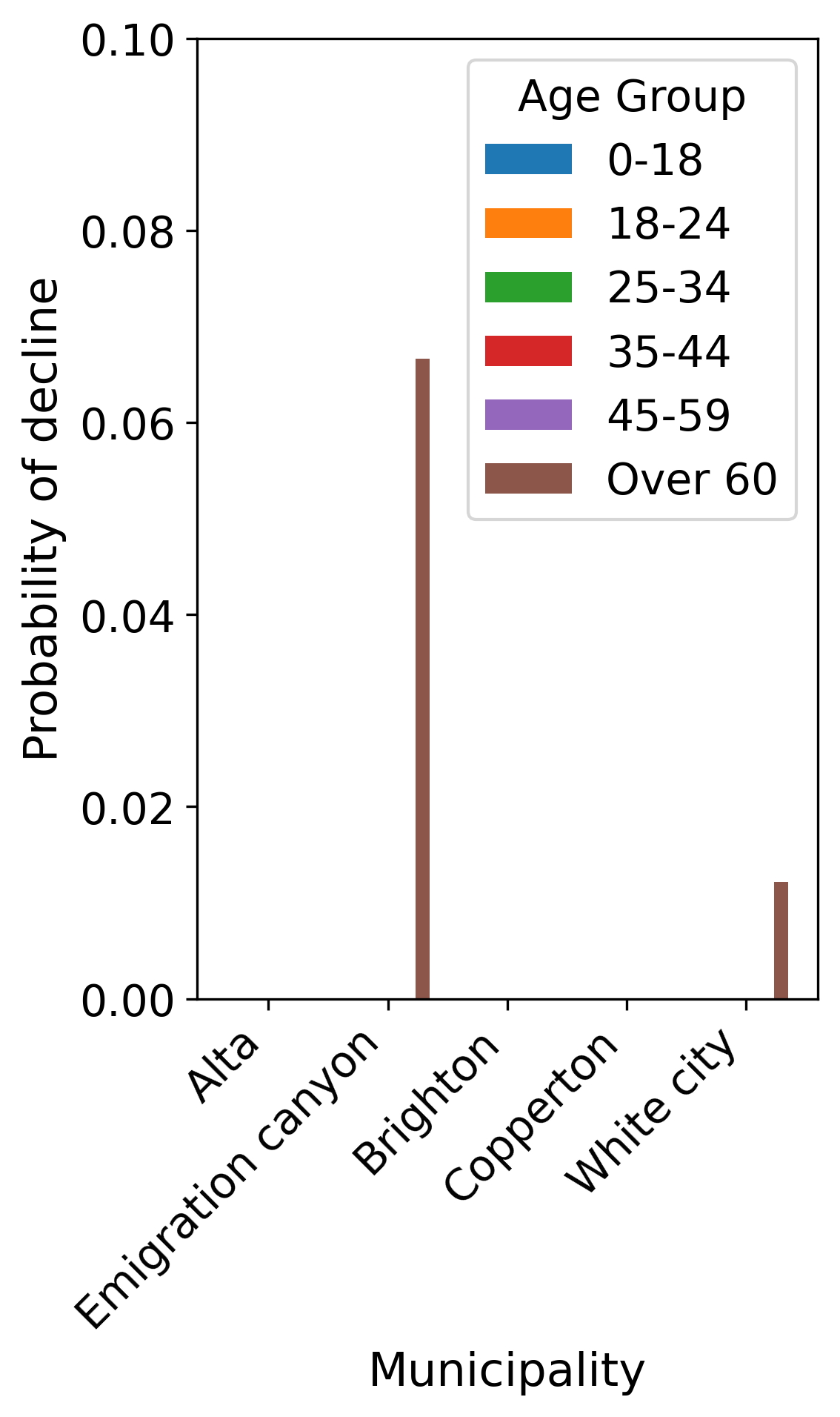}
\caption{LLM inference within the \framework framework across small municipalities. The $y$-axis shows the probability to decline participation in an outdoor activity.}
\label{fig:LLM_city_small}
\end{figure}

Small unincorporated or census-designated communities---Alta, Emigration Canyon, Brighton, Copperton, and White City---exhibit very small probabilities, mostly near zero. Only the over 60 age group shows some activity (Figure~\ref{fig:LLM_city_small}). It indicates that the model doesn't have enough information about these communities.

\section{Conclusion}\label{sec:con}
The individual-based ABM network models are well known for their utility in scenario analysis and for informing policy decisions.
On the other hand, LLMs are trained on massive amounts of text data that can be used to infer human behavior.
Therefore, leveraging both methods is an effective approach for modeling, in near real--time, a real-world problem.
To this end, we develop the \framework framework to strengthen individual-based temporal network modeling by reducing a data gap through LLM reasoning.
Notably, \framework is scalable and can model millions of agents.
 
LLMs contribute to modeling and simulation efforts across a range of domains and applications.
In this research, we use an LLM to predict human behavior.
Previous studies have assessed the performance of LLMs, particularly GPT-4, in predicting human decisions~\cite{xiao2025evaluating, peters2024large}.
These studies find that LLMs tend to reflect population-level patterns rather than individual decision-making, while also exhibiting biases that vary by context and group.
Moreover, LLMs have the potential to tailor their outputs so they can capture the nuances of specific countries or locations~\cite{qu2024performance}.
In this work, we use an LLM to predict human behavior at the group level rather than at the individual level.
We also provide contextual information about location and demographic characteristics.
Here, we use a zero-shot LLM inference to address previously unseen scenarios.
In future, we will explore RAG (retrieval-augmented generation) to improve the inference.
Additionally, we plan to include weather data to improve predictions of outdoor activity.

In the \framework framework, we address two data gaps in the individual-based network model: partially available activity data with unknown specific activity locations, and static activity data.
To address the first gap, we propose a systematic method to develop a temporal network model from activity data to leverage the available activity information while making realistic, data-dependent assumptions.
The temporal networks display degree patterns consistent with those observed in prior studies~\cite{del2007mixing}.

ABM uses activity or mobility data.
Unfortunately, collecting this activity data is a rigorous and time-consuming process; therefore, ABMs often use outdated or static activity data that do not reflect dynamic situations.
To address this second data gap, we incorporate an LLM component in the \framework framework.
In this approach, the ABM simulates disease spreading, and the LLM predicts the group-level human decision-making and behavior.
Overall, our results indicate that the epidemic curves generated by the \framework framework align more closely with observed data for Salt Lake County, UT than the curves produced by the ABM-only model.

The \framework framework opens the door to utilizing the powerful tool of LLM reasoning in network simulation for policy making, without replacing well-developed computational models, but instead assisting in addressing data gaps.



\enlargethispage{20pt}

\section*{Ethics}{This work did not require any ethical approval from a human subject or animal welfare committee.}

\section*{Data access}{The code and supplementary data are available in the public repository in Zenodo: \cite{moon_2026_19829913}.}


\section*{Competing}{We declare we have no competing interest.}

\section*{Funding}{This material is based upon work supported by the US Department of Energy, Office of Science, Office of Advanced Scientific Computing under Award Number DE-SC-ERKJ422.}

\section*{acknowledgment}{This manuscript has been authored by UT-Battelle LLC under contract DE-AC05-00OR22725 with the US Department of Energy.
The publisher acknowledges the US government license to provide public access under the DOE Public Access Plan (http://energy.gov/downloads/doe-public-access-plan).}



\printbibliography

@inproceedings{chopraagenttorch,
  title={AgentTorch: Agent-based Modeling with Automatic Differentiation},
  author={Chopra, Ayush and Subramanian, Jayakumar and Krishnamurthy, Balaji and Raskar, Ramesh},
  booktitle={Second Agent Learning in Open-Endedness Workshop}
}

@article{zhang2025llm,
  title={LLM-AIDSim: LLM-Enhanced Agent-Based Influence Diffusion Simulation in Social Networks},
  author={Zhang, Lan and Hu, Yuxuan and Li, Weihua and Bai, Quan and Nand, Parma},
  journal={Systems},
  volume={13},
  number={1},
  pages={29},
  year={2025},
  publisher={MDPI}
}

@article{vezhnevets2023generative,
  title={Generative agent-based modeling with actions grounded in physical, social, or digital space using Concordia},
  author={Vezhnevets, Alexander Sasha and Agapiou, John P and Aharon, Avia and Ziv, Ron and Matyas, Jayd and Du{\'e}{\~n}ez-Guzm{\'a}n, Edgar A and Cunningham, William A and Osindero, Simon and Karmon, Danny and Leibo, Joel Z},
  journal={arXiv preprint arXiv:2312.03664},
  year={2023}
}

@article{ghaffarzadegan2023generative,
  title={Generative agent-based modeling: Unveiling social system dynamics through coupling mechanistic models with generative artificial intelligence},
  author={Ghaffarzadegan, Navid and Majumdar, Aritra and Williams, Ross and Hosseinichimeh, Niyousha},
  journal={arXiv preprint arXiv:2309.11456},
  year={2023}
}

@inproceedings{ferraro2024agent,
  title={Agent-based modelling meets generative ai in social network simulations},
  author={Ferraro, Antonino and Galli, Antonio and La Gatta, Valerio and Postiglione, Marco and Orlando, Gian Marco and Russo, Diego and Riccio, Giuseppe and Romano, Antonio and Moscato, Vincenzo},
  booktitle={International Conference on Advances in Social Networks Analysis and Mining},
  pages={155--170},
  year={2024},
  organization={Springer}
}

@article{williams2023epidemic,
  title={Epidemic modeling with generative agents},
  author={Williams, Ross and Hosseinichimeh, Niyousha and Majumdar, Aritra and Ghaffarzadegan, Navid},
  journal={arXiv preprint arXiv:2307.04986},
  year={2023}
}

@misc{NHTS,
  author       = {Federal Highway Administration (FHWA) },
  title        = {National Household Travel Survey},
  howpublished = {\url{https://nhts.ornl.gov/}},
  year         = {2025},
  note         = {Accessed: 2025-08-21}
}

@misc{H3,
  author       = {Isaac Brodsky },
  title        = {H3: Uber’s Hexagonal Hierarchical Spatial Index},
  howpublished = {\url{https://www.uber.com/blog/h3/}},
  year         = {2025},
  note         = {Accessed: 2025-08-21}
}

@misc{NCES,
  author       = {National Center for Education Statistics},
  title        = {Average public school class size: Average class size in public K–12 schools, by school level, class type, and state: 2020–21},
  howpublished = {\url{https://nces.ed.gov/surveys/ntps/estable/table/ntps/ntps2021_sflt07_t1s}},
  year         = {2025},
  note         = {Accessed: 2025-06-01}
}

@misc{WSJ,
  author       = {Rachel Bachman},
  title        = {How Close Do You Need to Be to Your Gym?},
  howpublished = {\url{https://www.wsj.com/articles/how-close-do-you-need-to-be-to-your-gym-1490111186}},
  year         = {2024},
  note         = {Accessed: 2024-01-01}
}

@article{dubey2024llama,
  title={The llama 3 herd of models},
  author={Dubey, Abhimanyu and Jauhri, Abhinav and Pandey, Abhinav and Kadian, Abhishek and Al-Dahle, Ahmad and Letman, Aiesha and Mathur, Akhil and Schelten, Alan and Yang, Amy and Fan, Angela and others},
  journal={arXiv e-prints},
  pages={arXiv--2407},
  year={2024}
}

@article{liu2015beyond,
  title={Beyond neighborhood food environments: distance traveled to food establishments in 5 US cities, 2009--2011},
  author={Liu, Jodi L and Han, Bing and Cohen, Deborah A},
  journal={Preventing chronic disease},
  volume={12},
  pages={E126},
  year={2015}
}

@article{blumenberg2024decisions,
  title={Decisions \& distance: The relationship between child care access and child care travel},
  author={Blumenberg, Evelyn and Wander, Madeline and Yao, Zhiyuan},
  journal={Journal of transport geography},
  volume={114},
  pages={103756},
  year={2024},
  publisher={Elsevier}
}

@article{akinlotan2023travel,
  title={Travel for medical or dental care by race/ethnicity and rurality in the US: findings from the 2001, 2009 and 2017 National Household Travel Surveys},
  author={Akinlotan, Marvellous and Khodakarami, Nima and Primm, Kristin and Bolin, Jane and Ferdinand, Alva O},
  journal={Preventive Medicine Reports},
  volume={35},
  pages={102297},
  year={2023},
  publisher={Elsevier}
}

@article{del2007mixing,
  title={Mixing patterns between age groups in social networks},
  author={Del Valle, Sara Y and Hyman, James M and Hethcote, Herbert W and Eubank, Stephen G},
  journal={Social Networks},
  volume={29},
  number={4},
  pages={539--554},
  year={2007},
  publisher={Elsevier}
}

@article{tuccillo2023urbanpop,
  title={UrbanPop: A spatial microsimulation framework for exploring demographic influences on human dynamics},
  author={Tuccillo, Joseph and Stewart, Robert and Rose, Amy and Trombley, Nathan and Moehl, Jessica and Nagle, Nicholas and Bhaduri, Budhendra},
  journal={Applied Geography},
  volume={151},
  pages={102844},
  year={2023},
  publisher={Elsevier}
}

@article{moon2021contact,
  title={Contact tracing evaluation for COVID-19 transmission in the different movement levels of a rural college town in the USA},
  author={Moon, Sifat A and Scoglio, Caterina M},
  journal={Scientific reports},
  volume={11},
  number={1},
  pages={4891},
  year={2021},
  publisher={Nature Publishing Group UK London}
}

@article{mundt2009long,
  title={Long-distance dispersal and accelerating waves of disease: empirical relationships},
  author={Mundt, Christopher C and Sackett, Kathryn E and Wallace, LaRae D and Cowger, Christina and Dudley, Joseph P},
  journal={The American Naturalist},
  volume={173},
  number={4},
  pages={456--466},
  year={2009},
  publisher={The University of Chicago Press}
}

@article{moon2019spatio,
  title={A spatio-temporal individual-based network framework for West Nile virus in the USA: Spreading pattern of West Nile virus},
  author={Moon, Sifat A and Cohnstaedt, Lee W and McVey, D Scott and Scoglio, Caterina M},
  journal={PLoS computational biology},
  volume={15},
  number={3},
  pages={e1006875},
  year={2019},
  publisher={Public Library of Science San Francisco, CA USA}
}

@article{yang2024baseline,
  title={A baseline structure inventory with critical attribution for the US and its territories},
  author={Yang, Hsiuhan Lexie and Laverdiere, Melanie and Hauser, Taylor and Swan, Benjamin and Schmidt, Erik and Moehl, Jessica and Reith, Andrew and Adams, Daniel and Morris, Bennett and McKee, Jacob and others},
  journal={Scientific Data},
  volume={11},
  number={1},
  pages={502},
  year={2024},
  publisher={Nature Publishing Group UK London}
}

@techreport{tuccillo2024downscaling,
  title={Downscaling Synthetic Populations to Realistic Residential Locations},
  author={Tuccillo, Joe},
  year={2024},
  institution={Oak Ridge National Laboratory (ORNL), Oak Ridge, TN (United States)}
}

@article{ozik2021population,
  title={A population data-driven workflow for COVID-19 modeling and learning},
  author={Ozik, Jonathan and Wozniak, Justin M and Collier, Nicholson and Macal, Charles M and Binois, Micka{\"e}l},
  journal={The International Journal of High Performance Computing Applications},
  volume={35},
  number={5},
  pages={483--499},
  year={2021},
  publisher={SAGE Publications Sage UK: London, England}
}

@inproceedings{spannaus2025data,
  title={Data Assimilation for Robust UQ Within Agent-Based Simulation on HPC Systems},
  author={Spannaus, Adam and Moon, Sifat Afroj and Gounley, John and Hanson, Heidi A},
  booktitle={Proceedings of the Platform for Advanced Scientific Computing Conference},
  pages={1--11},
  year={2025}
}

@article{civitarese2025large,
  title={Large language models are zero-shot recognizers for activities of daily living},
  author={Civitarese, Gabriele and Fiori, Michele and Choudhary, Priyankar and Bettini, Claudio},
  journal={ACM Transactions on Intelligent Systems and Technology},
  volume={16},
  number={4},
  pages={1--32},
  year={2025},
  publisher={ACM New York, NY}
}

@inproceedings{kwon2023efficient,
  title={Efficient memory management for large language model serving with pagedattention},
  author={Kwon, Woosuk and Li, Zhuohan and Zhuang, Siyuan and Sheng, Ying and Zheng, Lianmin and Yu, Cody Hao and Gonzalez, Joseph and Zhang, Hao and Stoica, Ion},
  booktitle={Proceedings of the 29th symposium on operating systems principles},
  pages={611--626},
  year={2023}
}

@article{geng2025jsonschemabench,
  title={Jsonschemabench: A rigorous benchmark of structured outputs for language models},
  author={Geng, Saibo and Cooper, Hudson and Moskal, Micha{\l} and Jenkins, Samuel and Berman, Julian and Ranchin, Nathan and West, Robert and Horvitz, Eric and Nori, Harsha},
  journal={arXiv preprint arXiv:2501.10868},
  year={2025}
}

@article{willard2023efficient,
  title={Efficient Guided Generation for Large Language Models},
  author={Willard, Brandon T and Louf, R{\'e}mi},
  journal={arXiv preprint arXiv:2307.09702},
  year={2023}
}

@article{liu2022effective,
  title={The effective reproductive number of the Omicron variant of SARS-CoV-2 is several times relative to Delta},
  author={Liu, Ying and Rockl{\"o}v, Joacim},
  journal={Journal of travel medicine},
  volume={29},
  number={3},
  pages={taac037},
  year={2022},
  publisher={Oxford University Press}
}

@article{ukhsa2023covid,
  title={COVID-19 Omicron variant infectious period and transmission from people with asymptomatic compared with symptomatic infection: A rapid review},
  author={UKHSA},
  journal={GOV-14430.$\{$UK Health Security Agency$\}$},
  year={2023}
}

@article{wang2023asymptomatic,
  title={Asymptomatic SARS-CoV-2 infection by age: a global systematic review and meta-analysis},
  author={Wang, Bing and Andraweera, Prabha and Elliott, Salenna and Mohammed, Hassen and Lassi, Zohra and Twigger, Ashley and Borgas, Chloe and Gunasekera, Shehani and Ladhani, Shamez and Marshall, Helen Siobhan},
  journal={The Pediatric Infectious Disease Journal},
  volume={42},
  number={3},
  pages={232--239},
  year={2023},
  publisher={LWW}
}

@article{subramanian2021quantifying,
  title={Quantifying asymptomatic infection and transmission of COVID-19 in New York City using observed cases, serology, and testing capacity},
  author={Subramanian, Rahul and He, Qixin and Pascual, Mercedes},
  journal={Proceedings of the National Academy of Sciences},
  volume={118},
  number={9},
  pages={e2019716118},
  year={2021},
  publisher={National Academy of Sciences}
}

@article{qu2024performance,
  title={Performance and biases of large language models in public opinion simulation},
  author={Qu, Yao and Wang, Jue},
  journal={Humanities and Social Sciences Communications},
  volume={11},
  number={1},
  pages={1--13},
  year={2024},
  publisher={Palgrave}
}

@article{xiao2025evaluating,
  title={Evaluating the ability of large Language models to predict human social decisions},
  author={Xiao, Feng and Wang, XT XiaoTian},
  journal={Scientific Reports},
  volume={15},
  number={1},
  pages={32290},
  year={2025},
  publisher={Nature Publishing Group UK London}
}

@article{peters2024large,
  title={Large language models can infer psychological dispositions of social media users},
  author={Peters, Heinrich and Matz, Sandra C},
  journal={PNAS nexus},
  volume={3},
  number={6},
  pages={pgae231},
  year={2024},
  publisher={Oxford University Press US}
}

@article{liu2025agentic,
  title={Agentic AI for sustainable development: Leveraging large language model-enhanced agent-based modeling for complex policy strategies},
  author={Liu, Jia’an and Chu, Chu and Zhao, Yilin and Aoki, Goshi and Xiao, Zhiqing},
  journal={Emerging Media},
  volume={3},
  number={3},
  pages={401--413},
  year={2025},
  publisher={SAGE Publications Sage UK: London, England}
}

@article{kwok2024utilizing,
  title={Utilizing large language models in infectious disease transmission modelling for public health preparedness},
  author={Kwok, Kin On and Huynh, Tom and Wei, Wan In and Wong, Samuel YS and Riley, Steven and Tang, Arthur},
  journal={Computational and Structural Biotechnology Journal},
  volume={23},
  pages={3254--3257},
  year={2024},
  publisher={Elsevier}
}

@article{zhang2025exploring,
  title={Exploring the role of large language models in the scientific method: from hypothesis to discovery},
  author={Zhang, Yanbo and Khan, Sumeer A and Mahmud, Adnan and Yang, Huck and Lavin, Alexander and Levin, Michael and Frey, Jeremy and Dunnmon, Jared and Evans, James and Bundy, Alan and others},
  journal={npj Artificial Intelligence},
  volume={1},
  number={1},
  pages={14},
  year={2025},
  publisher={Nature Publishing Group UK London}
}

@article {Cramer2021-hub-dataset,
	author = {Cramer, Estee Y and Huang, Yuxin and Wang, Yijin and Ray, Evan L and Cornell, Matthew and Bracher, Johannes and Brennen, Andrea and Castro Rivadeneira, Alvaro J and Gerding, Aaron and House, Katie and Jayawardena, Dasuni and Kanji, Abdul H and Khandelwal, Ayush and Le, Khoa and Niemi, Jarad and Stark, Ariane and Shah, Apurv and Wattanachit, Nutcha and Zorn, Martha W and Reich, Nicholas G and US COVID-19 Forecast Hub Consortium},
	title = {The United States COVID-19 Forecast Hub dataset},
	year = {2021},
	doi = {10.1101/2021.11.04.21265886},
	URL = {https://www.medrxiv.org/content/10.1101/2021.11.04.21265886v1},
	journal = {medRxiv}
}

@article{afroj2023all,
  title={Are all underimmunized measles clusters equally critical?},
  author={Afroj Moon, Sifat and Marathe, Achla and Vullikanti, Anil},
  journal={Royal Society Open Science},
  volume={10},
  number={8},
  year={2023},
  publisher={The Royal Society}
}

@misc{moon_2026_19829913,
  author       = {Moon, Sifat Afroj},
  title        = {HALE data and simulation},
  month        = apr,
  year         = 2026,
  publisher    = {Zenodo},
  doi          = {10.5281/zenodo.19829913},
  url          = {https://doi.org/10.5281/zenodo.19829913},
}

\end{document}